\documentclass[runningheads]{llncs}
\usepackage{graphicx}
\usepackage{amssymb}
\usepackage{multirow}
\usepackage{bm}
\usepackage{booktabs}
\usepackage{floatrow}
\usepackage[font=small,labelfont=bf]{caption}

\newfloatcommand{capbtabbox}{table}[][\FBwidth]

\begin{document}
\title{Ultrasound Video Summarization using Deep Reinforcement Learning}
\titlerunning{Ultrasound Video Summarization using Deep RL}
%

\author{Tianrui Liu,~Qingjie Meng,~Athanasios Vlontzos,~Jeremy Tan,~Daniel Rueckert and Bernhard Kainz}
%
\institute{Department of Computing, Imperial College London, United Kingdom}
\maketitle              
\begin{abstract}

Video is an essential imaging modality for diagnostics, e.g. in ultrasound imaging, for endoscopy, or movement assessment. However, video hasn't received a lot of attention in the medical image analysis community. In the clinical practice, it is challenging to utilise raw diagnostic video data efficiently as video data takes a long time to process, annotate or audit. In this paper we introduce a novel, fully automatic video summarization method that is tailored to the needs of medical video data. Our approach is framed as reinforcement learning problem and produces agents focusing on the preservation of important diagnostic information. We evaluate our method on videos from fetal ultrasound screening, where commonly 
only a small amount of the recorded data is used diagnostically. We show that our method is superior to alternative video summarization methods and that it preserves essential information required by clinical diagnostic standards.  

 

\keywords {Video summarization  \and Reinforcement learning \and Ultrasound diagnostic}
\end{abstract}

\section{Introduction}

Ultrasound is a popular modality in medical imaging because of its low cost, real-time capabilities, wide availability and safety. 
It's primary output is a video stream. 
It is challenging to utilize video data retrospectively since it often contains too much redundant information, is too large  for easy documentation or audit and complicates remote assessment. Hence, finding a way to summarise the data without losing important information is vital.

In this paper, we present an ultrasound imaging summarization method using deep reinforcement learning. We show  effectiveness for the example of fetal ultrasound screening. 
Given video captures from full examinations of 30 to 60 minutes per video, our goal is to select a small subset of frames to create a summary video that is much shorter but contains sufficient, dynamic and essential information to facilitate retrospective analysis. Our deep summarization network adopts an encoder-decoder convolutional neural network structure which first extracts visual features from frame sequence and then feeds these features into a bi-directional long short-term memory network (Bi-LSTM) for sequential modeling.
The reinforcement learning (RL) network interprets the summarization task as a decision making process and takes actions on whether a frame should be selected for the summary set or not. The RL network maximizes expected rewards computed on the quality of the selected frames in terms of their representativeness, diversity, as well as the likelihood of being a standard diagnostic view plane~\cite{FASP2015}. The proposed method can be trained in either a supervised or unsupervised way. Hence, in case the training process of our summarization network does not have clinical annotations, the proposed method can be trained in a fully unsupervised way which can still achieve encouraging performance. 

\textbf{Contribution:} the contribution of this paper is three-fold: (1) We discuss a deep RL-based framework for ultrasound video summarization. To the best of our knowledge, this is the first method to use RL for this task. (2) We propose a novel diagnostic view plane reward for the RL network which encourages agents to select essential clinical information. (3) We take fetal screening as an example from the clinical practice and show experimental evidence for the effectiveness of our approach.

\textbf{Related Work:}
There has been much work done in computer vision on general video summarization techniques. Early works adopt low-level or mid-level visual features to locate important segments of a video with a particular strategy such as clustering \cite{dynamic_cluster2013,SumMe_ECCV2014,Tianrui_vSumm2014} and sparse dictionary learning \cite{sparsedictionary2011,SparseLatentDictionary_2014_CVPR}. In \cite{LSTMvSumm2016}, long short-term memory (LSTM) has been used to model the frame-level features for video summarization.  
In~\cite{FCSN_eccv2018}, Rochan \textit{et al}. \cite{FCSN_eccv2018} demonstrated that it is possible to model the video summarization task as an element-wise segmentation problem using fully convolutional sequence networks (FCSN).
Recently, Zhou and Qiao~\cite{RLvSumm_AAAI2018} propose an RL-based deep network for general video summarization. They formulate the video summarization task as a sequential decision making process and generate video summaries by predicting the probabilities of a given frame being a key-frame. Video summarization fits the ideas of RL well. RL has become increasingly popular in medical imaging research due to its effectiveness for various tasks. For example, Alansary~\textit{et al}. \cite{vlontzos2019multiple,alansary2019evaluating} have shown that RL can be successfully used for landmark detection in medical image analysis.

In many existing video summarization methods \cite{vSumm_ECCV_2002,RLvSumm_AAAI2018,FCSN_eccv2018}, the ground-truth depends on subjective human perception of frame importance. This is more variable than defining factual image classification tasks. Different annotators may provide significantly different labels for the same video sequence. It usually requires more than ten human annotators to mitigate inter-observer variance. In medical image analysis we have the advantage that importance is often defined according to diagnostically decision criteria. In~\cite{hysteroscopy2012MIA}, a summarization approach for hysteroscopy data was proposed. Based on the motion estimation of the camera capturing the hysteroscopy videos, they make use of physicians' attention on video segments for data summarization.  In~\cite{mehmood2014video}, a video summarization-based tele-endoscopy service is introduced. 
They compute image moments, curvature, and multi-scale contrast to obtain the saliency map of each frame for key frames selection. In this paper we explore how to effectively exploit the prior knowledge from diagnostically decision criteria for ultrasound videos. 


\section{Method}
\begin{figure*}[t]
\begin{centering}
\center\includegraphics[width=1.0\columnwidth]{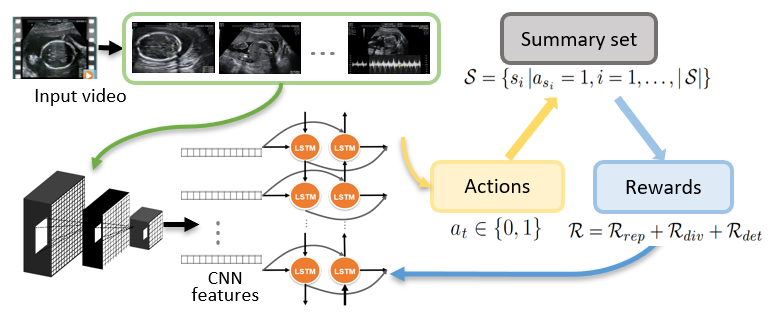}
\par\end{centering}
\caption{Overview of the proposed deep ultrasound video summarization network. Given an input video, a feature extraction network first computes CNN feature representations for frame sequences and feeds them into a bi-directional LSTM neural network. Then, the RL network takes actions $a_t$ on whether a frame should be selected for the summary set $\mathcal{S}$ or not. The RL network maximizes the expected rewards $\mathcal{R}$ computed on the quality of the selected frames in terms of their representativeness $\mathcal{R}_{rep}$, diversity $\mathcal{R}_{div}$, and likelihood of being a diagnostically valuable view plane $\mathcal{R}_{det}$.}
\label{fig:overview}
\end{figure*}
An overview over our RL-based ultrasound summarization network is illustrated in Fig.~\ref{fig:overview}.
Given an input video, a deep summarization network is used to extract deep feature representations from the input video sequence and sequentially models the frame features. It adopts an encoder-decoder convolutional network structure. 
Our encoder network is a diagnostic view plane detection network~\cite{sononet2017} pre-trained with ultrasound standard plane detection annotations. 
The decoder network takes the extracted feature maps of each input frame as input and feeds them into a Bi-LSTM to analyze features of both, past and future frames. 

Following the feature extraction, the RL network interprets the diagnostic video summarization task as a decision making process, in which a decision is to include a current frame in the summary or not. 
The RL network accepts latent scores from the Bi-LSTM as input and takes actions $a_t$ on whether a frame should be selected into the summary set $\mathcal{S}$ or not by maximizing the expected rewards $\mathcal{R}$. The rewards are computed on the quality of the selected frames in terms of their representativeness $\mathcal{R}_{rep}$, diversity $\mathcal{R}_{div}$, as well as the likelihood of being a standard diagnostic plane $\mathcal{R}_{det}$.

During training, the parameters of the decoder network will be learned using back-propagation, while the parameters of the encoder network are frozen.



Given the frame sequence $\{\bm{x}_t\}^T_{t=1}$, the outputs of the decoder network are frame-level probability scores, given by the sigmoid activation of the final fully connected layer, \emph{i.e.}, $p_t={sigmoid}(\bm{W}(\bm{x}_t,h_t^f,h_t^b),\bm{b})$
where $\bm{W}$ and $\bm{b}$ are the trainable parameters of the fully connected layer, and $h_t^f$ and $h_t^b$ denote the forward and backward hidden state of the input data $\bm{x}_t$. 
The frame selection agent takes an action $a_t$ according to these frame-level probability scores. In this work, the actions are defined as binary values, \emph{i.e.}, $a_t \in \{0,1\}$, indicating whether frame $t$ should be selected for the summary video or not. The frame selection is sampled by a Bernoulli distribution, \emph{i.e.}, $a_t \thicksim B(p_t)$.

\subsubsection{Reward Function}
In order to enable the deep summarization network to select a good set of key frames for the video summary, the deep summarization network maximizes three reward terms during training:
\begin{equation}
    \mathcal{R} = \mathcal{R}_{rep} + \mathcal{R}_{div} + \mathcal{R}_{det},
\end{equation}
where $\mathcal{R}_{det}$ evaluates the likelihood of a frame being a standard diagnostic plane, $\mathcal{R}_{rep}$ defines the representativeness reward  and the diversity reward $\mathcal{R}_{div}$ evaluates the quality of the selected summary $\mathcal{S}=\left\{s_{i}\left|a_{s_{i}}=1, i=1, \ldots,\right| \mathcal{S} |\right\}$ in terms of their representativeness and diversity, respectively.

The representativeness reward $\mathcal{R}_{rep}$ is defined as 
\begin{equation}
    \mathcal{R}_{\mathrm{rep}}=\exp (-\frac{1}{T} \sum_{t=1}^{T} \min _{t^{\prime} \in \mathcal{S}}\left\|\bm{x}_{t}-\bm{x}_{t^{\prime}}\right\|_{2}).
\end{equation}

It measures how well the generated summary can represent the original video by minimizing the mean squared errors (MSE) between video frames and their nearest medoids. Maximizing $\mathcal{R}_{rep}$ can therefore help to preserve the temporal information across the entire diagnostic video.

The diversity reward $\mathcal{R}_{\mathrm{div}}$ measures the dissimilarity between the selected frames of the summary video:
\begin{equation}
    \mathcal{R}_{\mathrm{div}}=\frac{1}{|\mathcal{S}|(|\mathcal{S}|-1)} \sum_{t \in \mathcal{S}} \sum_{i \in \mathcal{S}, i \neq t} d\left(\bm{x}_{t}, \bm{x}_{i}\right),
\end{equation}
where $d(\cdot , \cdot)$ measures the cosine dissimilarity of two vectors.

We further propose a novel standard plane detection reward term $\mathcal{R}_{\mathrm{det}}$ to encourage the agent to select essential diagnostic information. 
\begin{equation}
    \mathcal{R}_{det}=\frac{1}{|\mathcal{S}|}\sum_{i \in \mathcal{S}} s_i \left(\delta({y}_i=1)-\delta({y}_i=0) \right),
\end{equation}
where $\delta(\cdot)$ is a Dirac delta function, $s_{i}$ is the standard plane detection score of the $i$-th frame in the summary set $\mathcal{S}$ resulting from the encoder network topped up with a softmax layer. $y_i=1$ indicates that the $i$-th frame in $\mathcal{S}$ is classified as standard diagnostic view plane.


\begin{figure*}[t]
\begin{centering}
\center\includegraphics[width=0.9\columnwidth]{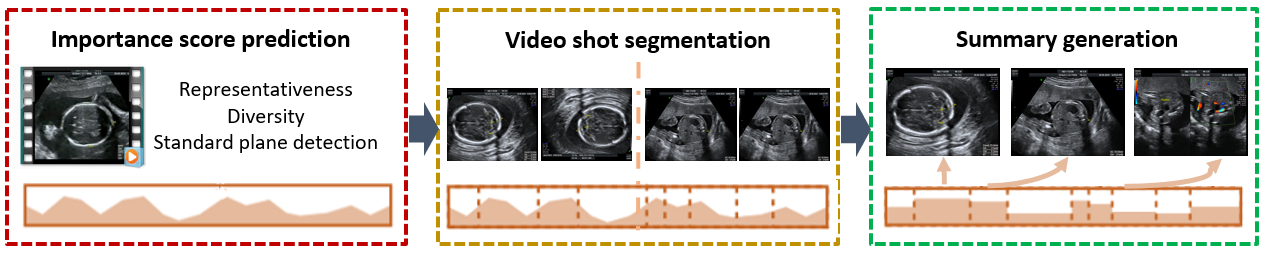}
\par\end{centering}
\caption{Sequential summary of ultrasound video summary generation modules.}
\label{fig:Vsumm_generation_precedure}
\vspace{-3mm}
\end{figure*}
\subsubsection{Optimization}

The learning objective is to train a video summarization agent for an optimal policy $\pi$ which indicates actions to take to maximize the overall reward. The expected reward $\mathcal{J}(\theta)$ is defined as $\mathcal{J}(\theta)= \mathbb{E}_{p_{\theta}\left(a|\pi\right)} (\mathcal{R})$
where $p_{\theta}\left(a|\pi\right)$ denotes the probability distribution over the actions of sequences. 

The proposed ultrasound video summarization method can be trained in either a supervised or an unsupervised way. For supervised training, we utilize a loss term that promotes to minimize the MSE between the predicted frame-wise importance scores and ground-truth scores, \emph{i.e.}, \begin{equation}
    \mathcal{L}_{pred}=\frac{1}{T} \sum_{t=1}^{T} \left\| p_{t}-p_{t}^*\right\|^{2}, 
\end{equation}
where $p_{t}$ and $p_{t}^*$ are the predicted frame-wise importance scores and the ground-truth user annotation scores, respectively.

We further propose to apply a regularization term to penalize the selection of a large number of frames in the summary set $\mathcal{L}_{reg} = || \frac{1}{T} \sum_{t=1}^{T}  p_{t}-\epsilon ||^{2}$ \label{eq:regloss} 
where $\epsilon$ is a scalar controlling the proportion of the selected frames and $\lambda$ controls the relative importance of the two loss terms. 

In our summarization network, the loss terms and reward terms are jointly optimized in an end-to-end manner. Thus, the total cost for the video summarization network is formulated as $\mathcal{L}_{sup}=\mathcal{L}_{pred} + \beta \mathcal{L}_{reg} - \gamma \mathcal{R}$ where $\beta$ and $\gamma$ are parameters to control the relevant importance of the costs.
In case that ground truth key frames through clinical annotations are limited or unavailable, our method can be used as fully unsupervised model by jointly optimizing the regularization and reward terms, \emph{i.e.}, $\mathcal{L}_{unsup} = \beta \mathcal{L}_{reg} - \gamma \mathcal{R}$. 


\subsubsection{Video Summary Generalization} 
Once we get the frame-level importance scores via the deep RL network, we can generate the video summary as given in Fig.~ \ref{fig:overview}. First, the input video is segmented into \emph{shots} using Kernel temporal segmentation (KTS) \cite{KTS_2014}. Then, we generate the video summaries by selecting the shots with the highest scores while keeping the duration of summaries below a threshold (\emph{e.g.}, 15$\%$ duration of the original video). The importance score of an shot equals to the average score of the frames in that shot.


\section{Experiments}

\subsubsection{Data}
In this paper, we use screen capture video recordings from fetal screening ultrasound examinations. There are 50 videos of 13-65 minutes length in our dataset from 50 different patients acquired between 24-30 weeks of gestation. 
The videos have been acquired and labelled during routine screenings according to the guidelines in the UK National Health Service (NHS) FASP handbook~\cite{FASP2015}. 
The feature extraction network is trained on annotations indicating the type of standard ultrasound diagnostic plane. From all available FASP planes we have selected Brain (Cb.), Brain (Tv.), Profile, Lips, Abdominal, Kidneys, Femur, Spine (Cor.), Spine (Sag.), 4CH, 3VV, RVOT, LVOT as the most frequent exemplars.


For ultrasound video summarization, we take the freeze-frame images which are saved by the sonongraphers during the scan as the ground-truth key frames. We follow the steps in \cite{LSTMvSumm2016,vSumm_NIPS2014} to convert key frame annotations into frame-level scores. The videos are temporally segmented into disjoint intervals using KTS \cite{KTS_2014}. If an interval (\emph{i.e.}, a shot) contains at least one key frame, we take this shot as a key shot and mark all the frames of it with score 1 and otherwise 0.
\vspace{-2mm}
\subsubsection{Evaluation Metrics}

Following the protocols proposed for general video summarization methods~\cite{Gygli_vSumm2015,TVSUM,vSumm_NIPS2014}, we compute the precision ($P$) and recall ($R$) according to the temporal overlap between a user annotated summary $\mathcal{A}$ and a predicted summary $\mathcal{B}$, \emph{i.e.}, $P =\frac{\mathcal{A} \cup \mathcal{B}}{|A|}$ and $R =\frac{\mathcal{A} \cup \mathcal{B}}{|B|}$,
where $|\mathcal{A}|$ denotes the duration of a summary $\mathcal{A}$ and $|\mathcal{A} \cup \mathcal{B}|$ is the temporal overlap between them two. The harmonic mean F1-score against is $F=2P \times R/(P+R)$.

\vspace{-2mm}
\subsubsection{Implementation}
The proposed video summarization approach is implementations in PyTorch. For the RL algorithm, the number of episodes of the episodic reinforcement learning algorithm is fixed to $5$. Stochastic Gradient Descent with momentum ($\rho$ = 0.9) and a weight decay of $1e^{-5}$ is used to train the models. The initial learning rate is set to $1e^{-4}$ and subsequently reduced by a factor of $0.5$ for every 50 epochs. The maximum training epoch is set as $300$. We set $\lambda = 0.01$ and $\epsilon =0.5$ for Eq.~\ref{eq:regloss}. $\beta$ and $\gamma$ are set as $0.01$ and 1. The experiments are performed on a single TITAN RTX GPU.

\subsection{Experimental Results}

\label{sec:SoA}

\begin{figure}[tp]
\begin{floatrow}
\capbtabbox{%
\begin{tabular}{ l  c } 

\toprule
Methods         & F1-scores\\ 
\midrule
FCSN$_{sup}$~\cite{FCSN_eccv2018} & 59.17 \\ 
DR-DSN$_{sup}$~\cite{RLvSumm_AAAI2018}  & 60.34   \\ 
Proposed$_{sup}$(our) & \textbf{63.29}\\ 
\midrule
 
DR-DSN$_{unsup}$~\cite{RLvSumm_AAAI2018}  & 40.92  \\ 
Proposed$_{unsup}$(our)    &   \textbf{56.73}\\ 
\midrule
\end{tabular}
}{%
 \vspace{3mm}
  \caption{Comparison to state-of-the-art video summarization methods. Best results in bold.}%
  \label{tab:SoA_comp}
}
\qquad
\ffigbox{%
 \includegraphics[height=3.5cm]{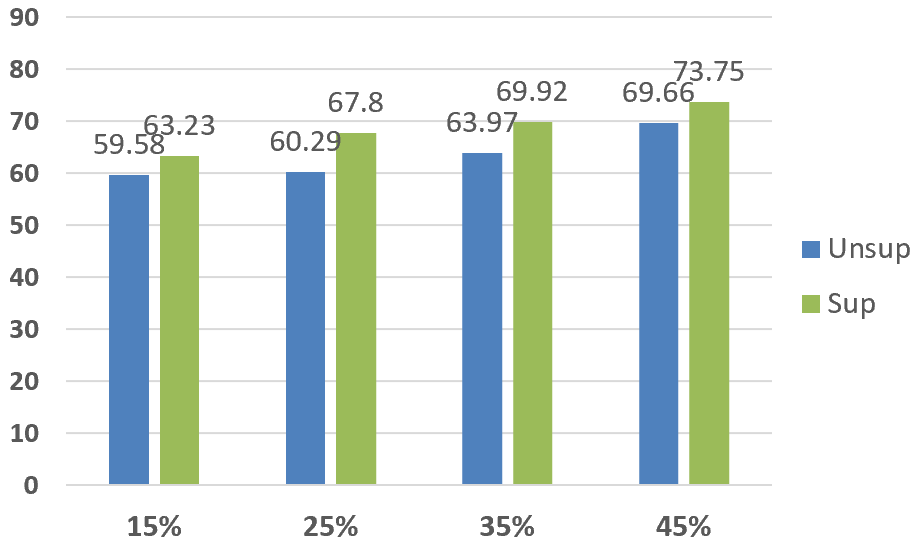}%
}
{%
  \caption{Performance of supervised vs. unsupervised summarization for different summary lengths.}%
  \label{fig:proportion}
}
\end{floatrow}
 \vspace{-3mm}
\end{figure}

We conduct experiments and compare with conventional state-of-the-art video summarization methods from traditional computer vision literature, including FCSN~\cite{FCSN_eccv2018} and DR-DSN~\cite{RLvSumm_AAAI2018}. We report the results in Table~\ref{tab:SoA_comp}. For fair comparison, we use the same feature extraction network which has been fine-tuned with our ultrasound standard plane annotations for the FCSN and DR-DSN approaches. 

We performed experiments on five different splits of training and testing subsets of percentages $80\%$ and $20\%$, \emph{i.e.}, 40 videos for training and 10 videos for testing each time. The averaged F1 scores for both unsupervised and supervised learning paradigm are compared in Table~\ref{tab:SoA_comp}. As we can see, the proposed ultrasound video summarization approach novel can effectively exploit the prior knowledge from standard diagnostic plane, leads to significant improvements for the performance especially for unsupervised model.

\begin{table}[tp]
\center
\caption{Comparison of results using different combinations of reward terms.}
\setlength{\tabcolsep}{2.8mm}{
\begin{tabular}{lllllllll}
 \toprule
\multicolumn{3}{c}{Rewards}  & \multicolumn{6}{c}{Learning paradigm}    \\ \midrule
\multirow{2}{*}{$\mathcal{R}_{{rep}}$ } 
& \multirow{2}{*}{$\mathcal{R}_{{div}}$  } 
& \multirow{2}{*}{$\mathcal{R}_{{det}}$  } 
& \multicolumn{3}{c}{Supervised} & \multicolumn{3}{c}{Unsupervised} \\ \cmidrule{4-9} 
         &          &          & F   & P &  R    & F & P & R \\  \midrule
$\times$    & $\times$      & $\times$      
&   58.41   & 57.43     & 59.52     &    -   & -        & -       \\ 
\checkmark  & $\times$      & $\times$      
&   59.36   & 58.27     & 60.59     & 37.65  &  37.25  & 38.07    \\ 
$\times$    & \checkmark    & $\times$      
&   57.34   &  56.41    & 58.40     & 32.57  &  32.03 & 33.15    \\  
\checkmark  & \checkmark    & $\times$      
&   59.20   &  58.22    & 60.31     & 38.82  & 38.42  & 39.24    \\ \midrule
$\times$    & $\times$      & \checkmark    
&   59.95    & 58.94    &  61.10    & 44.28  & 43.48  & 45.18   \\ 
\checkmark  & $\times$      & \checkmark    
&   61.98   &  60.88    & 63.22     & 54.56 & 53.53  & 55.72    \\
$\times$    & \checkmark    & \checkmark    
&   62.33   & 61.21     & 63.60     & 49.23 & 48.35  & 50.22    \\ 
\checkmark  & \checkmark    & \checkmark    
&   63.23   &  62.08    & 64.54     & 59.58  & 58.56  & 60.74   \\ 
\bottomrule
\end{tabular}}
\label{tab:award_comp}
\vspace{-3mm}
\end{table}

\subsubsection{Ablation study about the effectiveness of rewards}
We conduct ablation studies to investigate the effectiveness of the reward terms. For all the experiments in this section, we keep the same split of training and testing video.

Table~\ref{tab:award_comp} reports experiments for the proposed approach with different combinations of reward terms. As this table shows, As we can see, the proposed novel diagnostic view reward, \emph{i.e.}, $\mathcal{R}_{ {det}}$ for ultrasound video, leads to significant improvements for the summarization performance especially for unsupervised model. By using the standard plane detection reward $\mathcal{R}_{ {det}}$ alone, the summarization performance can be as good as $59.95$ regarding F1-score for the supervised model and $44.28$ for the unsupervised model. Compared to the results of unsupervised learning using  $\mathcal{R}_{ {div}}$ and $\mathcal{R}_{ {rep}}$ rewards on their own, the F1-score improves 11.45 and 6.25, respectively. For supervised learning, the improvements are 2.61 and 0.6. Using the combination of all the three reward terms leads to the highest scores, which are 63.23 and 59.58 for the supervised and the unsupervised model, respectively.

\subsubsection{Results with different summary lengths}
The above experiments are conducted with a constant summary length constraint of $15\%$, \emph{i.e.}, the length of the summary videos are restricted to be shorter than $15\%$ of the input video length. We also perform experiments on summaries generated with four different summary length constraints: $15\%$, $25\%$, $35\%$ and $45\%$ and show the results in Fig.~\ref{fig:proportion}. When the video summarization network is allowed to select more key shots, the F1 scores increase for both supervised models (green bars) and unsupervised models (blue bars).

\subsubsection{Qualitative Results}

\begin{figure*}[t]
\begin{centering}
\center\includegraphics[width=1.00\columnwidth]{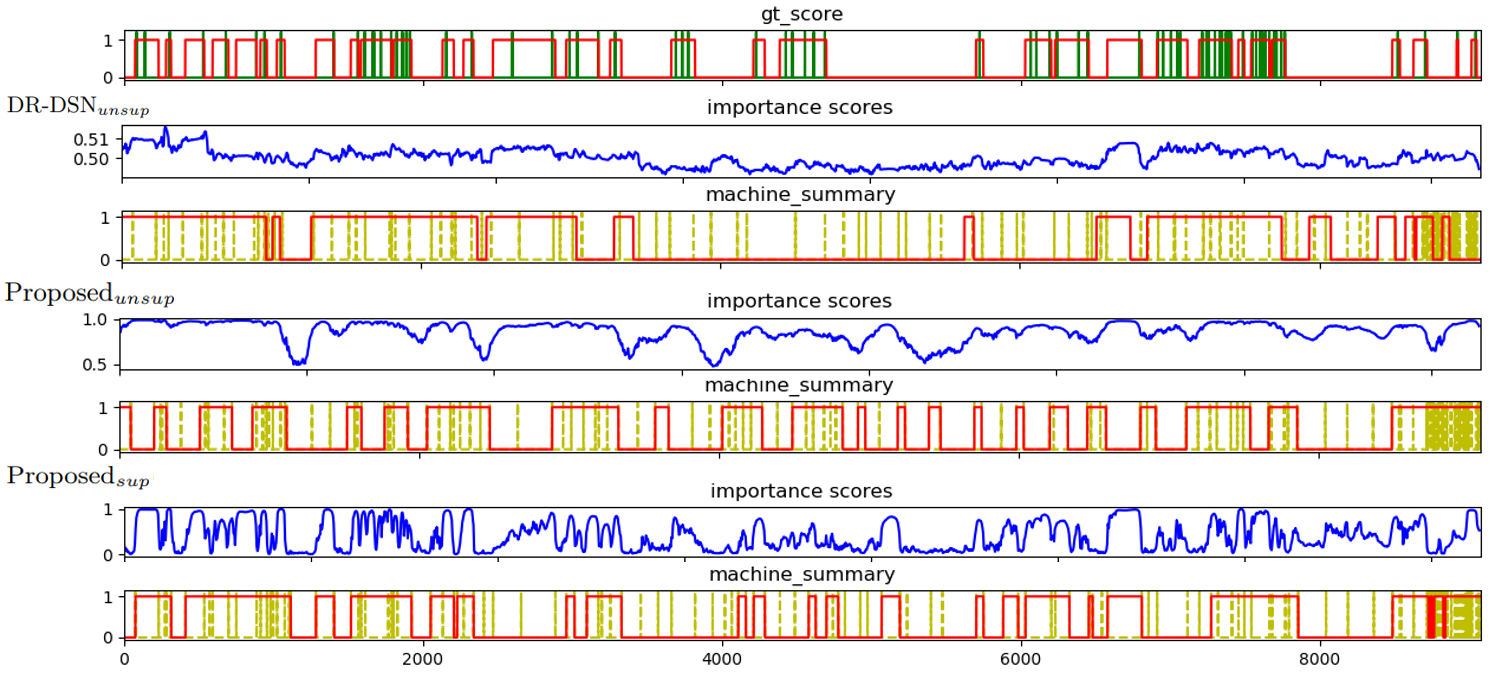}
\par\end{centering}
\caption{Comparison of video summary result \textbf{}on exemplar ultrasound video.}
\label{fig:Egkeyframes_vSumm}
\vspace{-2mm}
\end{figure*}

Fig.~\ref{fig:Egkeyframes_vSumm} shows a visualization of an example video summary generated by the proposed method and a baseline method ~\cite{RLvSumm_AAAI2018}. The ground-truth (gt) summary is shown at the top, where the gt key frames and gt scores are shown in green and red, respectively. We observe that the summary result using our approach have higher percentage of overlap with the gt. This implies that our method is able to preserve essential information for generating optimal and meaningful summaries.

\section{Conclusion}
We have proposed an RL-based deep learning model for effective diagnostic video summarization. 
The proposed framework has the potential to save storage costs as well as to increase efficiency when browsing patient video data during retrospective analysis or audit without loosing essential information. 
Both the supervised and unsupervised training model of our method can achieve good performance. Hence, our ultrasound video summarization method can be used for a variety of applications also when clinical annotations are unavailable. Experiments and ablation studies show that the proposed novel diagnostic view reward leads to significant improvements for the summarization performance and is able to summarize ultrasound videos without discarding important information. Future work will focus on experiments for other kind of diagnostic videos such as endoscopy videos or physiotherapeutic movement assessment videos.


\noindent\textbf{Acknowledgements:} 
We thank the volunteers and sonographers from routine fetal screening at St. Thomas' Hospital London. This work was supported by the Wellcome Trust IEH Award [102431] and EPSRC EP/S013687/1. The research was funded/supported by the National Institute for Health Research (NIHR) Biomedical Research Center based at Guy's and St Thomas' NHS Foundation Trust, King's College London and the NIHR Clinical Research Facility (CRF) at Guy's and St Thomas'. Data access only in line with the informed consent of the participants, subject to approval by the project ethics board and under a formal Data Sharing Agreement. The views expressed are those of the author(s) and not necessarily those of the NHS, the NIHR or the Department of Health. 

\bibliographystyle{splncs04}
\bibliography{ifind}

\end{document}